\documentclass[letterpaper]{article} 
\usepackage{aaai21}  
\usepackage{times}  
\usepackage{helvet} 
\usepackage{courier}  
\usepackage[hyphens]{url}  
\usepackage{graphicx} 
\usepackage{natbib}  
\usepackage{caption} 
\usepackage{subfigure}
\usepackage{flushend}
\usepackage{mathrsfs}
\usepackage{bbm}
\usepackage{booktabs}
\usepackage{amsthm,amsmath,amssymb}
\title{Recovering Surveillance Video Using RF Cues}
\author{
   Xiang Li \textsuperscript{\rm 1},
   Rabih Younes \textsuperscript{\rm 2},\\
}
\affiliations{
    \textsuperscript{\rm 1} Department of Electrical and Computer Engineering, Carnegie Mellon University, PA, USA. \\
    \textsuperscript{\rm 2} Department of Electrical and Computer Engineering, Duke University, NC, USA.
}

\begin{document}
\maketitle

\begin{abstract}
Video capture is the most extensively utilized human perception source due to its intuitively understandable nature. A desired video capture often requires multiple environmental conditions such as ample ambient-light, unobstructed space, and proper camera angle. In contrast, wireless measurements are more ubiquitous and have fewer environmental constraints. In this paper, we propose CSI2Video, a novel cross-modal method that leverages only WiFi signals from commercial devices and a source of human identity information to recover fine-grained surveillance video in a real-time manner. Specifically, two tailored deep neural networks are designed to conduct cross-modal mapping and video generation tasks respectively. We make use of an auto-encoder-based structure to extract pose features from WiFi frames. Afterward, both extracted pose features and identity information are merged to generate synthetic surveillance video. Our solution generates realistic surveillance videos without any expensive wireless equipment and has ubiquitous, cheap, and real-time characteristics.
\end{abstract}

\section{Introduction}
In the human perception field, camera-based and radio frequency (RF)-based approaches have been extensively researched and achieve substantial success in many tasks such as 2D \cite{alp2018densepose,sun2019deep,cao2017realtime} and 3D \cite{chen20173d,rayat2018exploiting,tome2017lifting} pose estimation and body segmentation \cite{wang2020solo,wang2020solov2,li2022r,li2022panoramic}. Nevertheless, both existing camera-based and RF-based approaches suffer from technical or practical issues. For instance, illumination and occlusion constrains for camera-based approaches, and equipment cost and energy consumption for LiDARs-based approaches \cite{wang2019person}. The latest works \cite{wang2019person,jiang2020towards,huang2021towards,huang2021forgery} prove that WiFi signals have the capability to achieve camera-based comparable human perception precision which can be further explored in many field. For example, surveillance, game and human robot interaction. However, since a WiFi signal is a 1D sequence, it cannot carry as much information as an 2D image. Limited by that, previous works never tried to recover the complete frames from WiFi signal. In this paper, we take a step forward to recover RGB surveillance video by merging WiFi signals and a weak source of the human identity profile.

\begin{figure}[t]
\includegraphics[width=0.49\columnwidth]{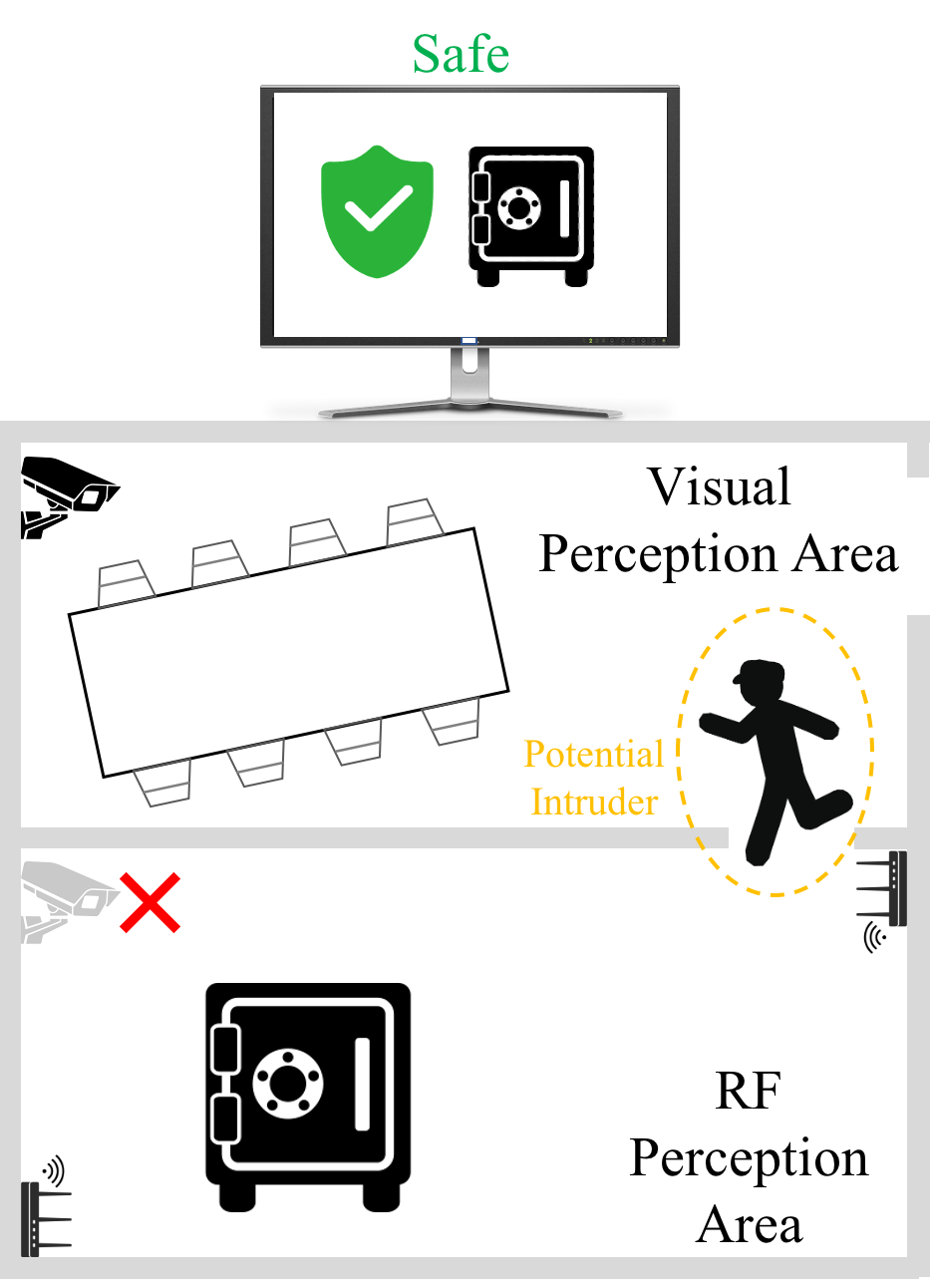}
\includegraphics[width=0.49\columnwidth]{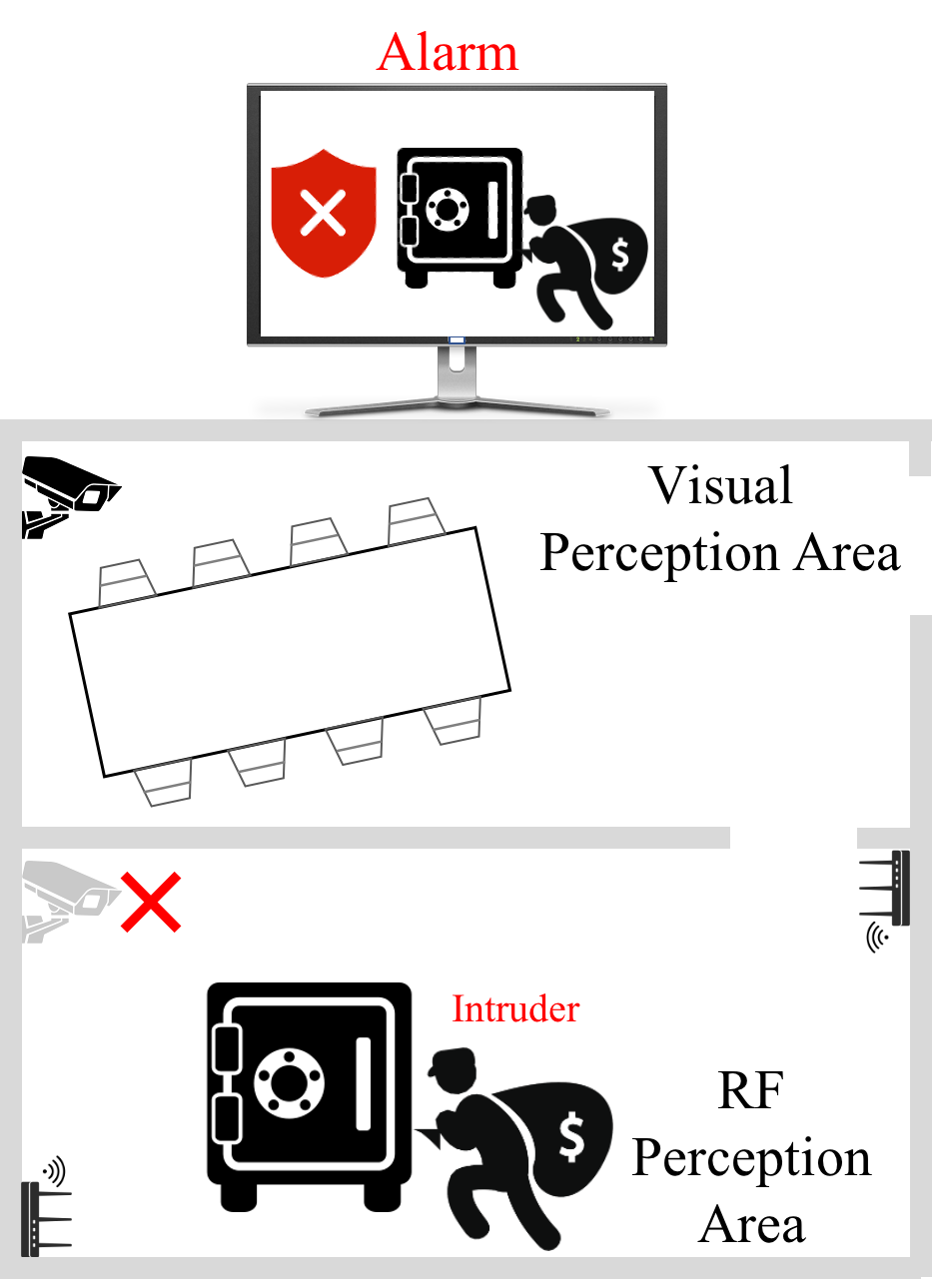}
\caption{An example of how the CSI2Video system works. The left is the first stage-recording of the identity of intruder. The right is the second stage-generating surveillance video after the intruder enters the RF perception area.}
\label{steps}
\end{figure}

We propose CSI2Video, an innovative cross-modal method that leverages only WiFi signals from commercial devices and a source of human identity information to recover fine-grained surveillance video in a real-time manner. The proposed CSI2Video is supposed to take WiFi signals as input and output colorful video frames which accurately reflect localization and movements of human instances. To make the system more applicable, the WiFi signal used in CSI2Video is designed to be collected from off-the-shelf WiFi routers. Thanks to the rapid development of internet in recent years, WiFi routers have been widely deployed in all areas of daily life. Thus, WiFi signal can be obtained without additional cost even out of lab. Moreover, in CSI2Video, we leverage the out-sourced human identity information to help the network to determine which instance to display on the screen. For example, those identity can be obtained by surveillance cameras in security scenario, user defined image for virtual portrait generation, or any kind of personal information from intelligent on-body devices. Because the source of identity profile has no constrains on our frameworks, in the rest of the paper, we will only show the results utilize video frames from surveillance video be the identity source from convenient consideration. Specifically, we assume the visual perception area and RF perception area are spatially and temporally separated since we only need to recognize human identity one time instead of tracking it. By separating the two perception areas, the CSI2Video system could work efficiently even when real-time surveillance video is unavailable (e.g. under attack or malfunction).

Figure~\ref{steps} shows an example of CSI2Video system which contains one surveillance camera and one WiFi router pair. The system works in a two-stage manner. In the first stage, the surveillance camera records the video in a normal way which aims to get the appearance profiles of people who enter the RF perception area. Then, in the second stage, the WiFi router pair record the wireless measurements which contain the direct localization and movement information then transmit the data to the central server. The center server conducts cross-modal mapping and video generation processes taking advantage of both real-time wireless measurements and formally recorded visual captures to produce the synthetic surveillance video.

Our contributions are summarized as follows. 
\begin{itemize}
\item To better visualize the spatial information encoded in WiFi signals and exploit the potential usage of WiFi signals in human perception field, we present a novel cross-modal video generation system, CSI2Video. To our best knowledge, this is the first attempt to generate realistic RGB surveillance video using WiFi signals in a real-time manner.
\item We propose a novel cross-modal mapping network that maps wireless measurements to human pose features.
\item We propose a novel video generation network taking pose features instead of keypoint coordinates as inputs. 
\item We propose a novel evaluation metric for RF-based video generation which can efficiently reflects both localization error and image quality of generated human figures. 
\end{itemize}

\section{Related Works}

\subsection{Human Perception}
\subsubsection{Camera-Based Human Perception}
Most works about human perception are rooted in the use of cameras that enables the well-developed feature extraction method in computer vision field. To date, several human perception tasks have achieved impressive performance and have been widely deployed in daily life such as cognitive process understanding \cite{li2020predicting}, pose estimation \cite{alp2018densepose,sun2019deep,cao2017realtime, chen20173d,rayat2018exploiting,tome2017lifting}, body segmentation \cite{wang2020solo,li2021video,li2022hybrid,li2022online,zhao2022self} as well as activity recognition \cite{ijjina2017human,garcia2018modality,kniaz2018thermalgan,luo2021toward,li2020activitygan}. More recently, cross-modal fusion is introduced to improve performance and avoid environmental constraints. For example, Zou Han et al \cite{zou2019wifi} proposed a cross-modal model which merges both visual and wireless features to conduct indoor activity recognition task and got almost perfect results. \cite{premebida2014pedestrian,matti2017combining,han2015real} combined LiDAR with RGB cameras for pedestrian detection. 

\subsubsection{RF-Based Human Perception}
The most popular sensing modalities for RF-based systems are 3D point clouds captured by LiDAR and depth map captured by Frequency Modulated Continuous Wave (FMCW). LiDAR-based sensing is a well-developed method in many human perception tasks such as human detection \cite{maturana2015voxnet}, tracking \cite{leigh2015person} etc.. The FMCW radar system was introduced by Adib et al\cite{adib2015capturing}. They first attempted to capture coarse human body using dedicated RF radios. After that, they continued to exploit the usability of FMCW radio for 2D \cite{zhao2018through} and 3D \cite{zhao2018rf} pose estimation through wall or occlusions. Compared with the above methods, which rely severely on dedicated equipment, WiFi signals provide a more ubiquitous and cheap solution. WiPose \cite{jiang2020towards} generated 3D human skeletons in a single person scenario. Person in WiFi \cite{wang2019person} proposed end-to-end body segmentation and two-stage pose estimation methods taking advantage of only commercial WiFi transceivers.

\subsubsection{On-Body Devices-Based Human Perception}
With the rapid advances in information transmission and battery storage, it has become much easier to deploy on-body devices. Compared to the fixed perception area of camera-based and RF-based approach, on-body devices provide a flexible solution. Since on-body sensor can be designed tiny and portable, it is convenient to be carried around. In the past, the main concern of on-body devices is battery storage capability, while battery-free and wireless charging technologies have greatly downplayed this problem. In addition, on-body devices can take a bunch of sources to monitor human body condition. For example, the most popular biosignal sources are ECG, EMG, EEG and motion signal sources are acceleration and angular velocity. By taking those signals, on-body devices has succeeded in human activity \cite{lara2012survey}, mood \cite{zenonos2016healthyoffice}, tachycardia\cite{acharya2017automated} etc. recognition.
 
\begin{figure*}[t]
\centering
\includegraphics[width=\linewidth]{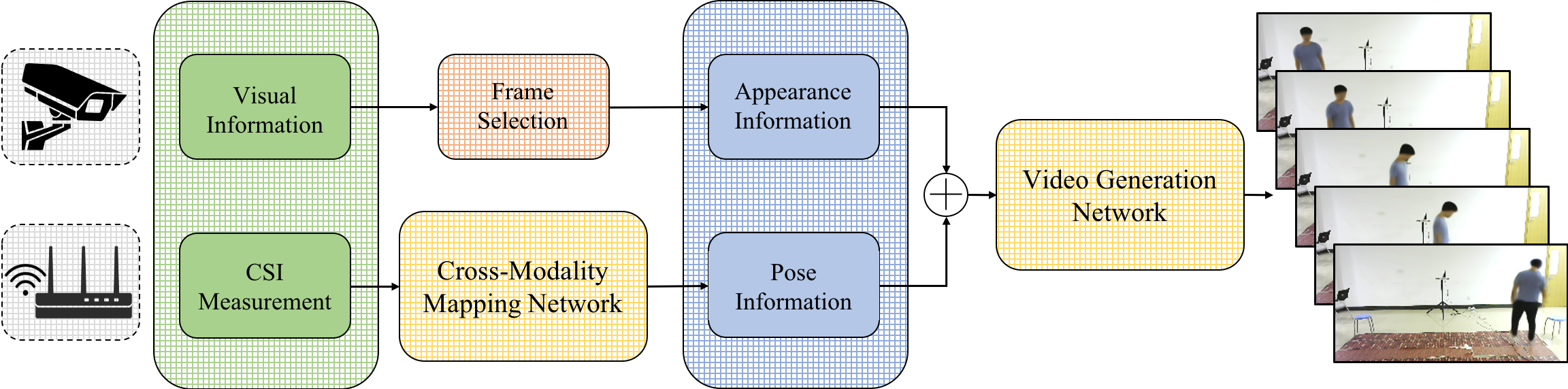}
\caption{System overview.}
\label{overview}
\end{figure*}

\subsection{Image \& Video Generation}
 Deep generative models have been extensively investigated in image or video generation tasks. Variational Auto-Encoders (VAEs) \cite{kingma2013auto} and Generative Adversarial Networks (GANs) \cite{goodfellow2014generative} are the most popular models. After the first debut of GANs, many follow-up works are committed to solving problems encountered in original GANs. Arjovsky et al noticed the model collapse and unstable training process in GANs and proposed an improved version wGAN \cite{arjovsky2017wasserstein} and wGAN-GP \cite{gulrajani2017improved}. \cite{mirza2014conditional} explored the performance of GANs in a conditional side. They first proposed cGAN to generate images controlled by class attributes. Besides, many other variants of GANs were proposed to improve the original GANs in different aspects. For example, DCGAN \cite{radford2015unsupervised} enables CNN to conduct feature extraction, StarGAN \cite{choi2018stargan} allows image-to-image transformation among multiple domains, PatchGAN \cite{isola2017image} introduces pixel-wise loss function. For human image generation, \cite{yang2018pose} proposed a pose-guided method to generate synthetic human video in a disentangled way-plausible motion prediction and coherent appearance generation. Similarly, Cai et al also proposed a two-stage human motion video generation method via GANs \cite{cai2018deep}. Though GANs is a powerful tool to generate videos and have made exceptional breakthrough in many tasks, we will not employ it in this work since CSI2Video system has a unique ground-truth. Thereby, the Mean Square Error (MSE) loss would not impair the image quality.

\section{CSI2Video}
\subsection{System Overview}
CSI2Video is a cross-modal system that merges both wireless and visual information to produce real-time surveillance video. Specifically, the visual and wireless measurements are gathered spatially and temporally separated to obtain appearance and localization features respectively. In our system, the wireless information is extracted from standard IEEE 802.11n WiFi signals using commercial devices. Due to the different electromagnetic properties of the background environment and human bodies, WiFi signals have embedded rich spatial information of human localization and movements. However, the WiFi signals transmit at the same time may not be received simultaneously by the antennas considering the different propagation paths resulted from the signal penetrates, refracts and reflects. Fortunately, in WiFi communication system, channel state information (CSI) was adopted to describe the current channel condition between each transceiver ($\mathit{T}$) and receiver ($\mathit{R}$) pairs. This way, the spatial information of human body is encoded into a $N_{tx}\times N_{rx}\times N_C$ matrix where $N_{tx}$ and $N_{rx}$ denote the numbers of transmitting and receiving antennas respectively and $N_C$ denotes the number of orthogonal frequency division multiplexing (OFDM) subcarriers. Let $\mathbf{C}$ be the single CSI measurement where $\mathbf{C}\in\mathbb{C}^{N_{tx}\times N_{rx}\times N_C}$. For each continuous CSI measurements, the recorded CSI data can be represented as $\mathcal{C}=\left\{ \mathbf{C_0}\cdots \mathbf{C_i}\cdots \mathbf{C_n}\right\}$. For human identity profile, it is used to give the network side information to determine which human instance should be displayed since the wireless measurement contains no explicit identity information. We choose video captures as our source of identity profile in this work. For each continuous video captures, the recorded frames can be denoted as $\mathcal{F}=\left\{ \mathbf{F_0}\cdots \mathbf{F_i}\cdots \mathbf{F_n}\right\}$. Therein, $\mathbf{F}$ indicates one single frame where $\mathbf{F}\in \mathbb{R}^{H\times W\times \mathit{3}}$.

As depicted in Figure~\ref{overview}, CSI2Video consists of two parts-cross-modal mapping and video generation. The cross-modal mapping network takes CSI measurements as input and outputs Joint Heat Maps (JHMs) and Part Affinity Fields (PAFs) \cite{cao2017realtime}. The PAFs and JHMs will be concatenated by the selected video frames which contain human appearance features and then be feed into the video generation network. Finally, the video generation network merges both human pose features and appearance features then produces the synthetic surveillance video. In general, the CSI2Video converts the spatial information of the human body contained in the wireless signals into a form of ordinary RGB surveillance video.

\subsection{Cross-Modal Mapping}
\begin{figure}[t]
\centering
\includegraphics[width=\columnwidth]{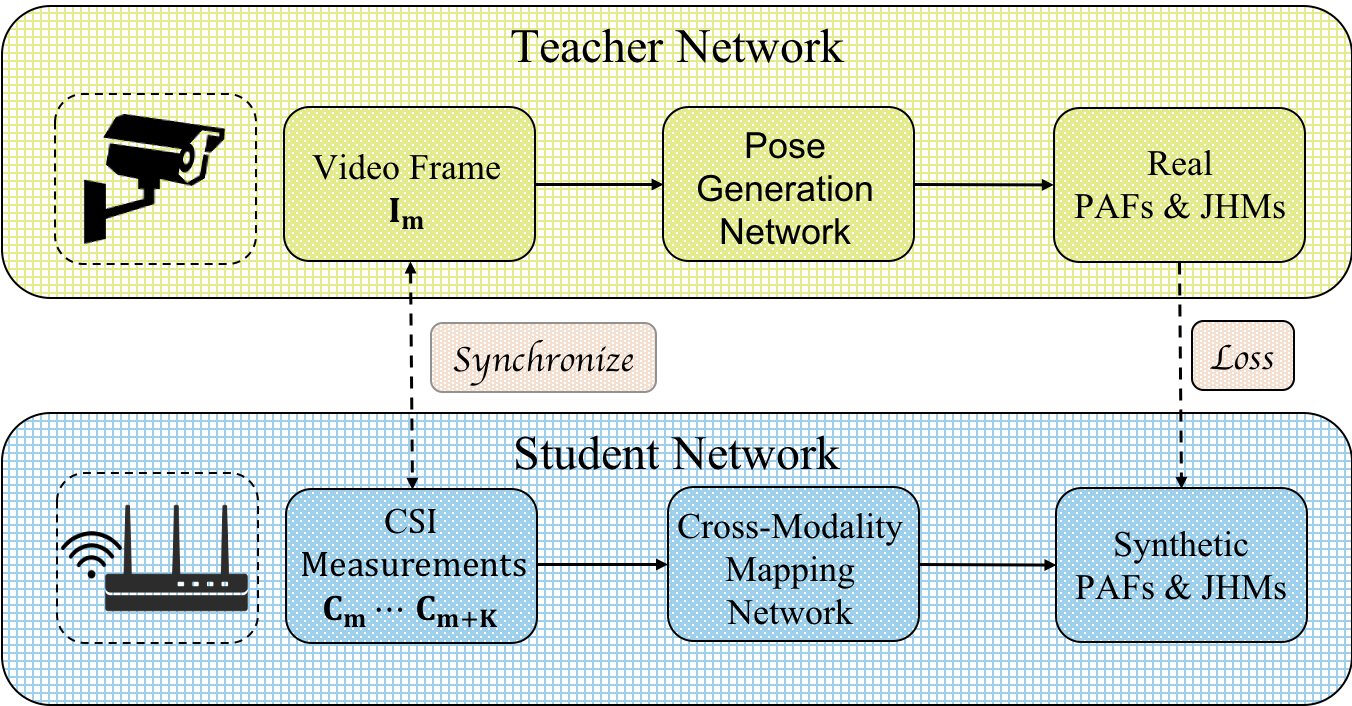}
\caption{Training phase of cross-modal mapping network.}
\label{csi2poseSystem}
\end{figure}
The cross-modal mapping network takes CSI measurements $\mathcal{C}$ as input and outputs PAFs and JHMs. So, to train the cross-modal mapping network, we need a set of synchronized CSI measurements $\mathcal{C}$ and their corresponding PAFs and JHMs pairs. For convenience, we employed Openpose \cite{cao2017realtime} to compute the PAFs \& JHMs for training phase and denote Openpose as $\mathcal{O}(\cdot)$. Let $\mathcal{I}=\left\{ \mathbf{I_0}\cdots \mathbf{I_i}\cdots \mathbf{I_n}\right\}$ be the frames used to compute body pose where $\mathbf{I}\in\mathbb{R}^{H\times W\times\mathit{3}}$ represents a single recorded frame. Since $\mathcal{C}$ is a 1D time-series with less points than $\mathcal{I}$, it can not contain as much information as $\mathcal{I}$. To ease the influence of unbalanced information density, we use a higher frequency to sample CSI measurement and pair multiple CSI measurements to one video frame. Let $K$ be number of CSI measurements which are corresponded to one video frame. Moreover, to feed a complex matrix to the neural network, we choose the amplitude of CSI measurement to represent the whole complex matrix. Therefore, the data pair used to train cross-modal mapping network can be represented as Equation~\ref{data} where $\mathbf{C_m}$ and $\mathbf{I_m}$ is assumed to have already been synchronized.
\begin{equation}
\mathbf{D}_m=\left\{(|\mathbf{C_m}|\cdots |\mathbf{C_{m+K}}|),\mathcal{O}(\mathbf{I_m})\right\}
\label{data}
\end{equation}

\subsection{Video Generation}
The video generation network takes two sources of input. The first source is generated PAFs ($\mathcal{P}$) and JHMs ($\mathcal{J}$) from the cross-modal mapping network which aims to reflect the localization and movement features. The second source is the visual captures for giving side information of human instances. Assuming $\mathcal{F}$ as a continuous video captures, however, we just need a few of them to give the video generation network cues about human appearances. Let $l$ be the number of selected frames from $\mathcal{F}$. Given PAFs $\mathbf{P_i}$, JHMs $\mathbf{J_i}$ and selected video frames $\mathcal{F}_l$, the video generation network outputs synthetic surveillance frames $\mathbf{S_i}$ as Equation~\ref{video generation}. Specifically, we interpolate all input to $d\times H_{\mathcal{G}}\times W_{\mathcal{G}}$ where $d$ is the number of channels of each source of input.
\begin{equation}
\mathbf{S_i}=\mathcal{G}(\mathbf{P_i}, \mathbf{J_i}, \mathcal{F}_l)
\label{video generation}
\end{equation}

\section{Experiment}
\subsection{Hardware Setup and Data Preparation}
\begin{figure}[t]
\centering
\includegraphics[width=0.49\linewidth, height=0.33\linewidth]{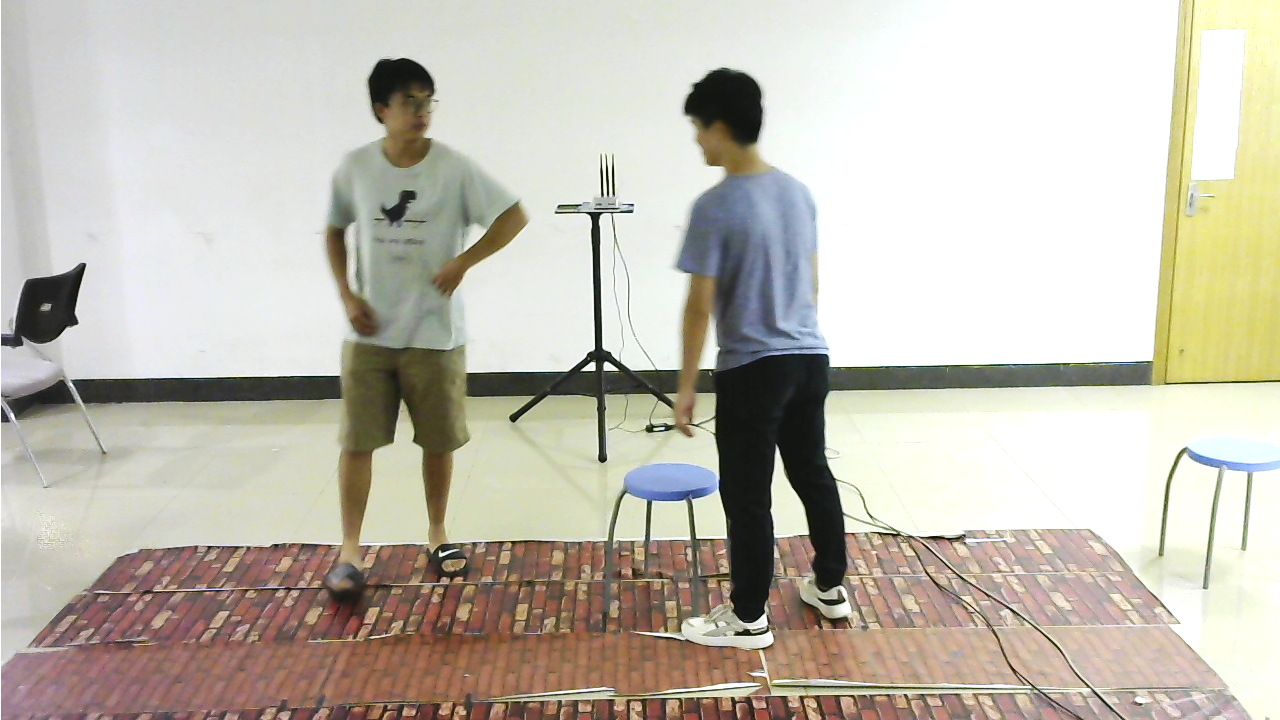}
\includegraphics[width=0.49\linewidth, height=0.33\linewidth]{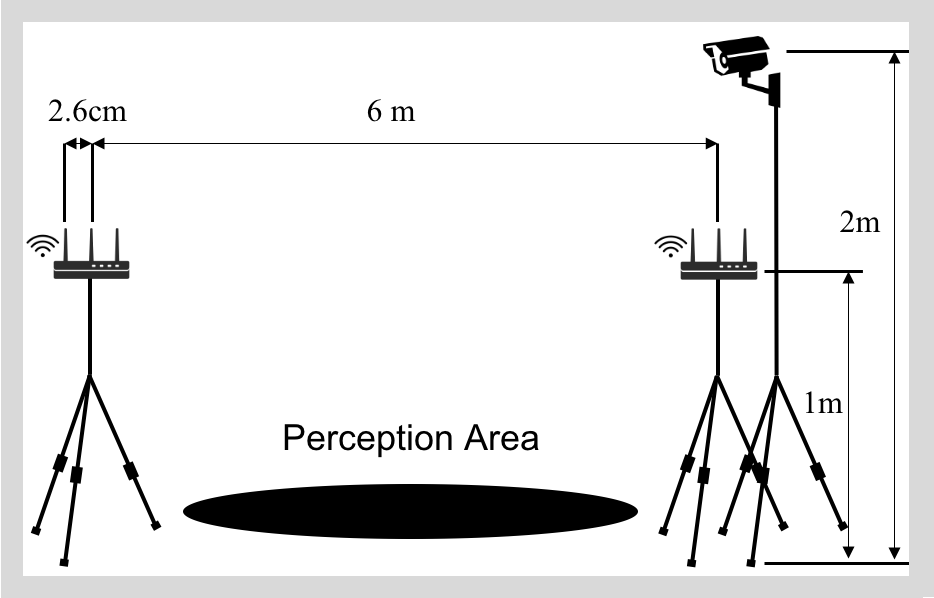}
\caption{The left is an example of captured video frame. The right is the sketch map of the setup of measurement equipment.}
\label{hardware}
\end{figure}

We collected the data in an $8m\times 16m$ office room with 5 volunteers. During the experiment phase, zero to three volunteers were asked to perform walking, sitting, waving hands and several random movements concurrently in the perception area. We leveraged one logitech 720p camera and two Intel 5300 NICs equipped with three antennas each to record the CSI and video frame pairs. As shown in Figure~\ref{hardware}, the NICs and RGB camera were placed at a height of 1m and 2m respectively. Antennas of each NICs were uniformly spaced at a distance of 2.6cm like common commercial WiFi routers and the two NICs were separated with each other about 6m away. Moreover, the camera was set to record $1280\times 720$p RGB video with a FPS of 7.5. For CSI measurement, we used an open-source tool to record \cite{halperin2011tool} and set the NICs to communicate with a bandwidth of 20 MHz centering in the 5.6 GHz WiFi band. In this condition, we recorded the CSI measurements of 30 subcarriers with a sampling rate of 100Hz.
Since the unstable nature of wireless communication, the sampling interval of the measured CSI signals are not strictly equal to 0.01s. Thus, before synchronizing the CSI measurements $\mathcal{C}$ and video frames $\mathcal{I}$, we conducted data cleaning to remove outliers and to resample the CSI measurements to exactly 100Hz. In our implementation, $K$ is set to 5. This means 5 CSI measurements are matched with one video frame. Consequently, we created 24082 pairs where 75\% of it is used for training and 25\% of it is used for testing. Specifically, to better fit our network structure, we interpolated the CSI measurements to an image-size tensor.

For video frames $\mathbf{F}_l$, we randomly selected them from $\mathcal{I}$ in our implementation for convenience consideration. Since we just suppose frame $\mathcal{F}$ contains human appearance information and take advantage of nothing beyond it, selecting $\mathbf{F}_l$ from $\mathcal{I}$ has no difference from using any other source and will not result in any negative influence on overall results. We set $l=1$ in our implementation.

\subsection{Networks}
In CSI2Video, two deep neural network structures are tailored. The cross-modal mapping network maps the CSI tensor to JHMs and PAFs and the video generation network makes use of JHMs, PAFs and video frames to generate synthetic surveillance video. Note that this is a proof-of-concept experiment that aims to show the ability to recover surveillance video from CSI measurements and can be further improved by carefully frame selection and manual annotations.

\subsubsection{Cross-Modal Mapping Network}
\begin{figure*}[t]
\centering
\includegraphics[width=\linewidth]{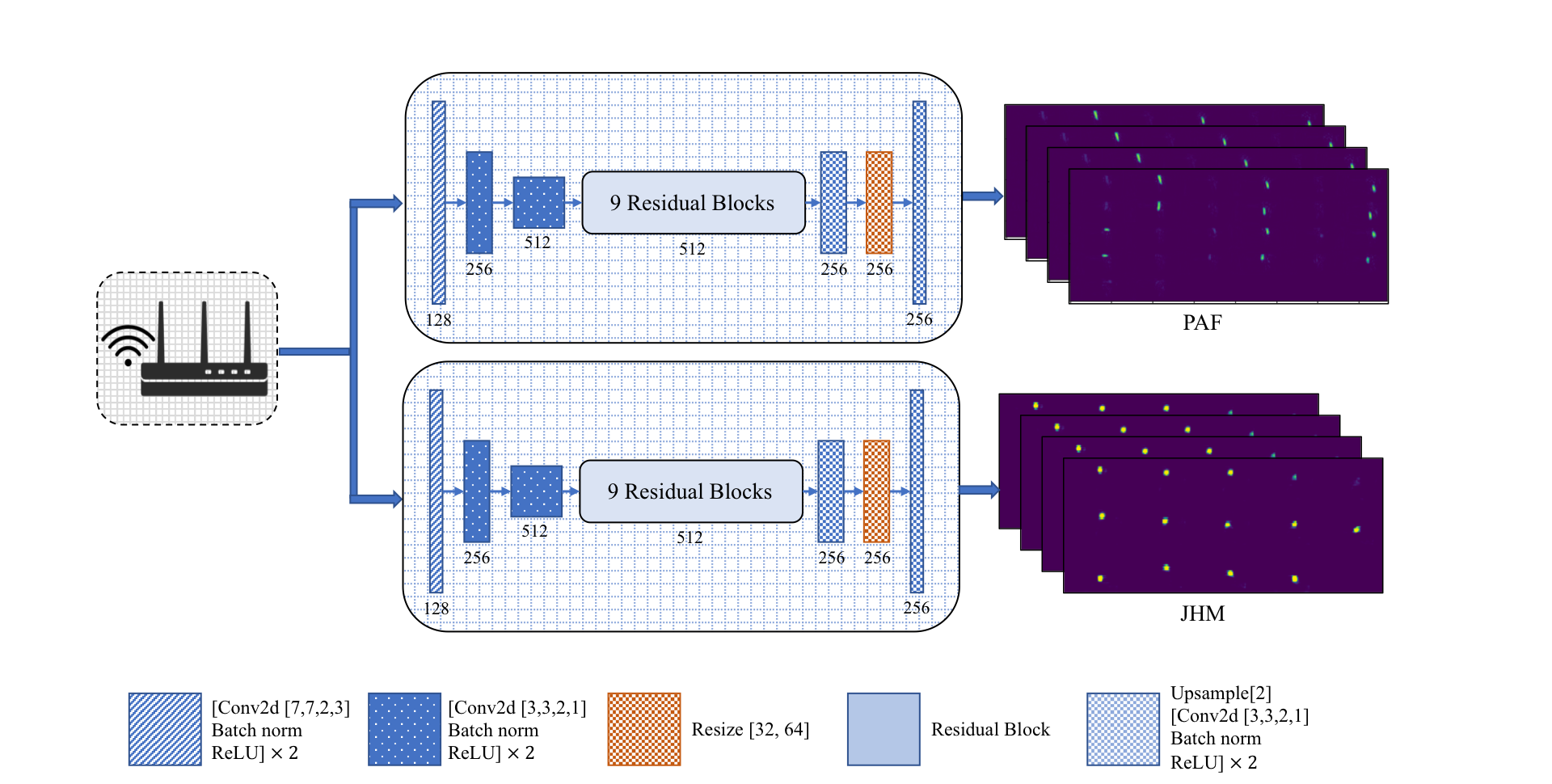}
\caption{Cross-modal mapping network}
\label{csi2pose}
\end{figure*}

Figure~\ref{csi2pose} shows the details of cross-modal mapping network. It is based on an auto-encoder structure and the embedded residual blocks \cite{he2016deep} serve for domain transformation. To obtain a larger receptive field, the first two layers have a $7\times 7$ kernel while other layers contain a $3\times 3$ kernel. In our implementation, we chose 14 keypoints of human body to force cross-modal mapping network to learn which are Nose, Neck, Rshoulder, RElbow, RWrist, LShoul- der, LElbow, LWrist, RHip, RKnee, LHip, LKnee, RAnkle, LAnkle. These 14 keypoints reflect the essential pose information to recover the human body in a video frame. Thereby, the output of the cross-modal mapping network should be $14\times H_J\times W_J$ for JHMs and $26\times H_P\times W_P$ for PAFs.

As depicted in Figure~\ref{csi2poseSystem}, we trained our cross-modal mapping network using a teacher-student architecture. By doing this, we obtain labels without laborious manual involvement. Moreover, we leveraged mean sum error (MSE) criterion and computed the loss function as Equation~\ref{loss}
\begin{equation}
\mathcal{L}_c=\lambda_J\mathcal{L_{\mathbf{J}}}+\lambda_P\mathcal{L_{\mathbf{P}}}
\label{loss}
\end{equation}
where $\mathcal{L_{\mathbf{J}}}$ and $\mathcal{L_{\mathbf{P}}}$ are losses on JHMs and PAFs respectively. $\lambda_J$ and $\lambda_P$ are scalar weights to balance the two losses of $\mathcal{L}_c$. Further, since JHMs and PAFs have almost zeros value in most points except for body parts, we added attention mechanism to force network coverage faster. The JHM loss $\mathcal{L}_J$ and pixel-wise weight $w^{c,h,w}_J$ can be computed as Equation~\ref{jhmloss} and Equation~\ref{jhmweight}.
\begin{equation}
\mathcal{L}_\mathbf{J}=||w_J^{c,h,w}\cdot(\hat{\mathbf{J}}^{c,h,w}-\mathbf{J}^{c,h,w})||_2^2
\label{jhmloss}
\end{equation}

\begin{equation}
w^{c,h,w}_J=\alpha_J\cdot|\hat{\mathbf{J}}^{c,h,w}|+\beta_J
\label{jhmweight}
\end{equation}
where $\alpha_J$ and $\beta_J$ are scalars to adjust the attention between JHMs and background, $\hat{\mathbf{J}}^{c,h,w}$ is the ground-truth of JHMs, $\mathbf{J}^{c,h,w}$ is the output of cross-modal network. Similarly, we computed the PAFs loss $\mathcal{L}_P$.

\begin{figure}[t]
\centering
\includegraphics[width=\linewidth]{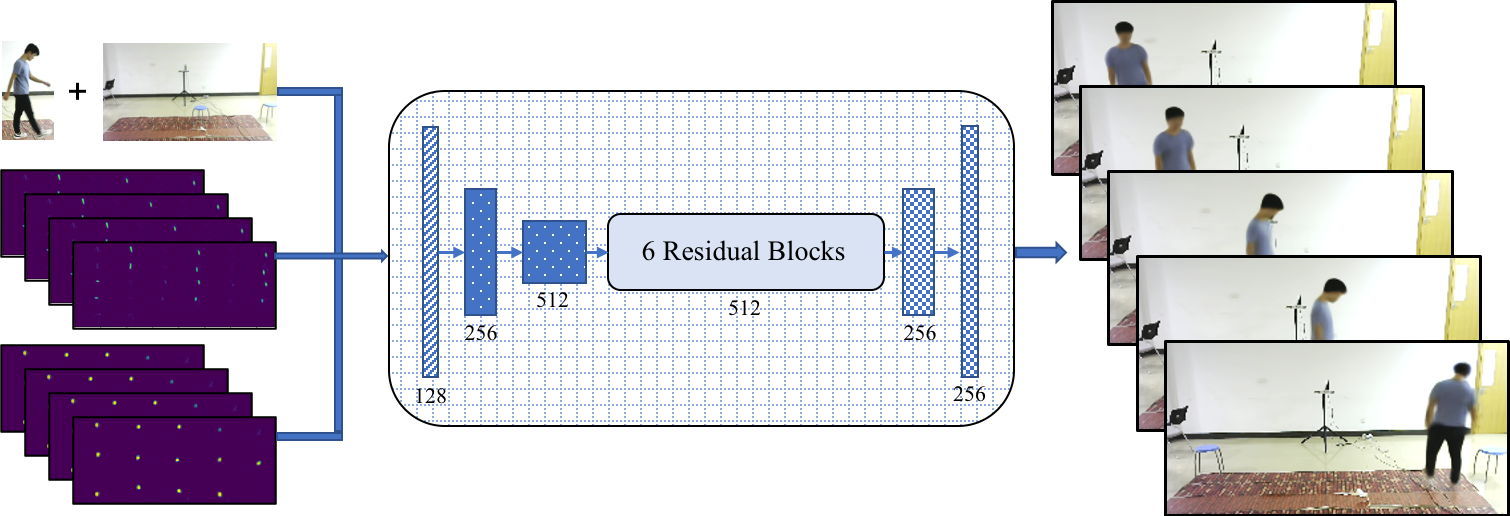}
\caption{The video generation network.}
\label{pose2video}
\end{figure}
\subsubsection{Video Generation Network}
Before feeding generated PAFs, JHMs and appearance profile to the video generation network, all inputs are interpolated to $d_{\mathbf{P,J},\mathcal{F}}\times H_{\mathcal{G}}\times W_{\mathcal{G}}$. The video generation network inherits the auto-encoder structure of cross-modal mapping network but with less residual blocks. In addition, unlike the cross-modal mapping network, the video generation network has a symmetry structure. Thus, the output frame is a $3\times H_{\mathcal{G}}\times W_{\mathcal{G}}$ tensor. We set $H_{\mathcal{G}}=128, W_{\mathcal{G}}=256$ in our implementation.

Since we trained the network on a fixed background, we need the network to pay more attention to the human body reconstruction task. For this reason, we utilized Mask-RCNN which can be denote as $\mathcal{M}(\cdot)$ to generate human masks to segment the foreground and background in order to compute the loss separately. The loss functions can be computed as Equation~\ref{fore} and Equation~\ref{back} where $\mathbf{B}^{c,h,w}$ is the selected background image. 
\begin{equation}
\mathcal{L}_{f}=\mathcal{M}(\mathbf{I}^{c,h,w})\times||\mathbf{S}^{c,h,w}-\mathbf{I}^{c,h,w}||_2^2
\label{fore}
\end{equation}
\begin{equation}
\label{back}
\mathcal{L}_{b}=(\mathbbm{1}^{c,h,w}-\mathcal{M}(\mathbf{I}^{c,h,w}))\times||\mathbf{S}^{c,h,w}-\mathbf{B}^{c,h,w}||_2^2
\end{equation}

Consequently, the total loss of video generation network can be computed as Equation~\ref{vloss}. Therein, $\lambda_f$ and $\lambda_b$ are scalars to balance the two losses.
\begin{equation}
\mathcal{L}_v=\lambda_f\mathcal{L}_f+\lambda_b\mathcal{L}_b
\label{vloss}
\end{equation}

\begin{figure*}[t]
\centering

\begin{minipage}[b]{1\linewidth}
\centering
\includegraphics[width=0.19\linewidth, height=0.11\linewidth]{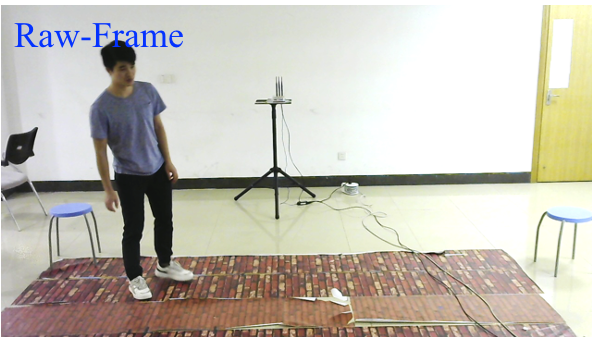}
\includegraphics[width=0.19\linewidth, height=0.11\linewidth]{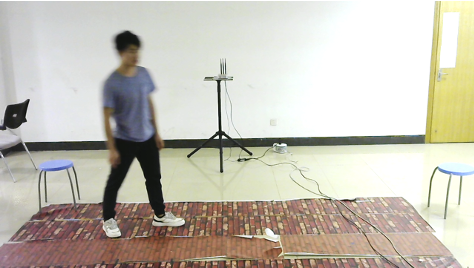}
\includegraphics[width=0.19\linewidth, height=0.11\linewidth]{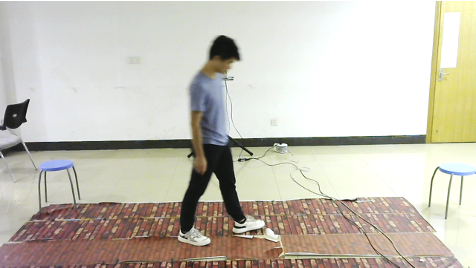}
\includegraphics[width=0.19\linewidth, height=0.11\linewidth]{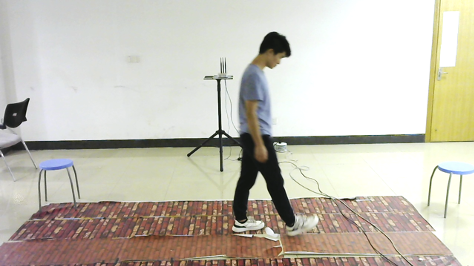}
\includegraphics[width=0.19\linewidth, height=0.11\linewidth]{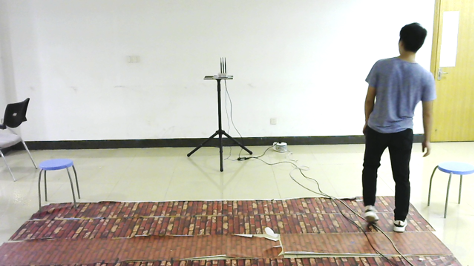}
\end{minipage}

\begin{minipage}[b]{1\linewidth}
\centering
\includegraphics[width=0.19\linewidth, height=0.11\linewidth]{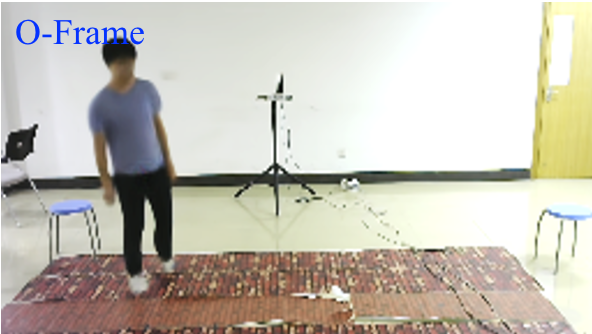}
\includegraphics[width=0.19\linewidth, height=0.11\linewidth]{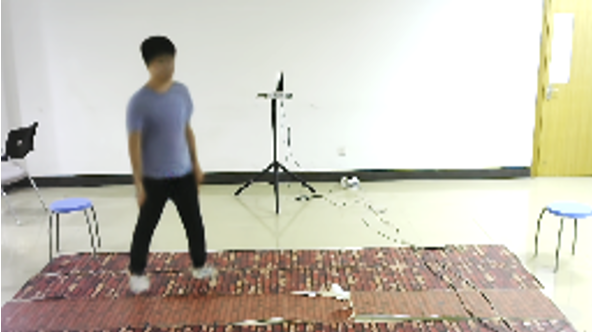}
\includegraphics[width=0.19\linewidth, height=0.11\linewidth]{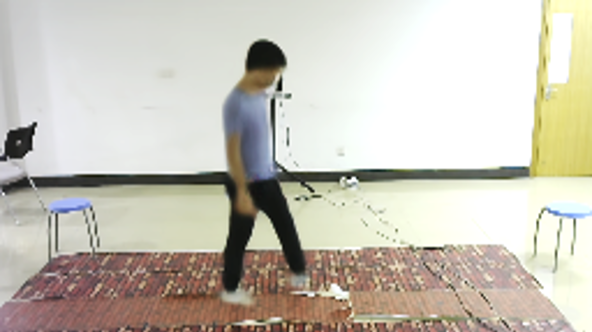}
\includegraphics[width=0.19\linewidth, height=0.11\linewidth]{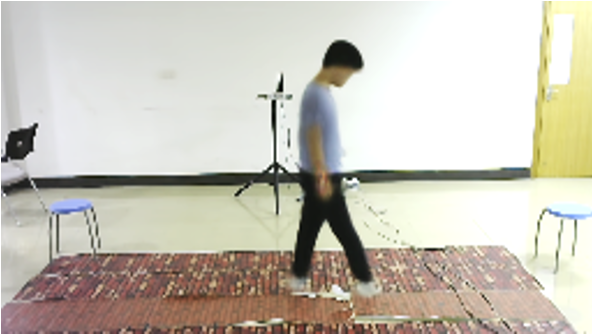}
\includegraphics[width=0.19\linewidth, height=0.11\linewidth]{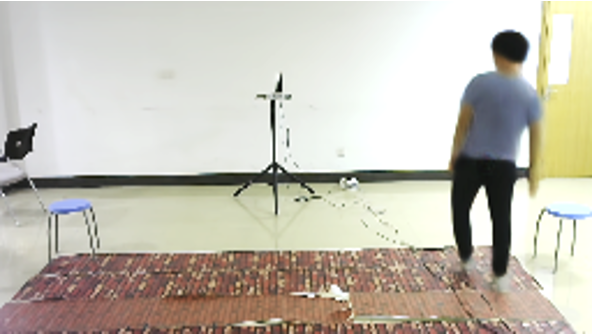}
\end{minipage}

\begin{minipage}[b]{1\linewidth}
\centering
\includegraphics[width=0.19\linewidth, height=0.11\linewidth]{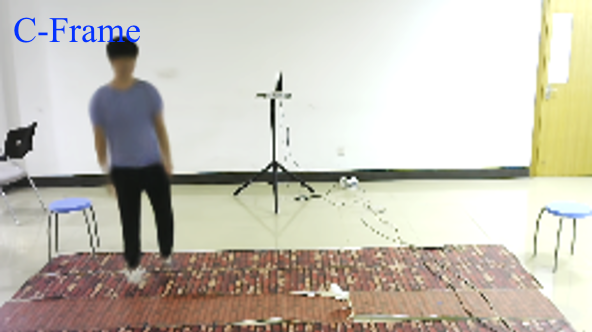}
\includegraphics[width=0.19\linewidth, height=0.11\linewidth]{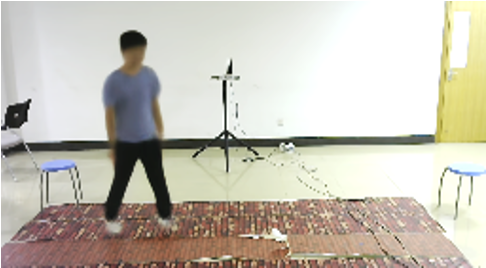}
\includegraphics[width=0.19\linewidth, height=0.11\linewidth]{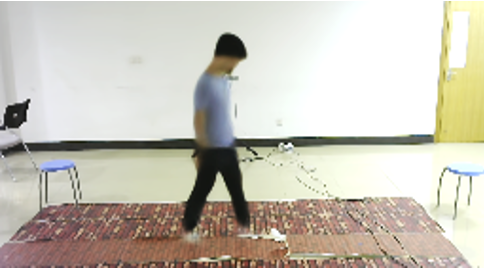}
\includegraphics[width=0.19\linewidth, height=0.11\linewidth]{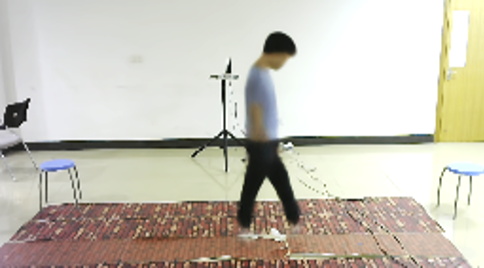}
\includegraphics[width=0.19\linewidth, height=0.11\linewidth]{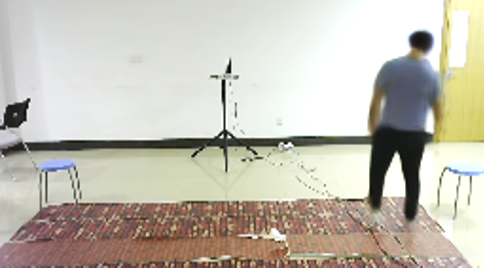}
\end{minipage}

\caption{Video generation results in single person scenario. The first row is the frames captured by camera. The second and third row are the frames generated from video generation network using PAFs and JHMs from Openpose and cross-modal mapping network respectively.}
\label{sp}
\end{figure*}
\subsection{Implementation Details}
We train each of the networks for 20 epochs on the Pytorch framework. An Adam optimizer with $\beta_1=0.9$ and $\beta_2=0.999$ is adapted in all of the training process. The learning rate starts from 1e-6 for cross-modal mapping network and from 1e-3 for video generation network. They are divided by 10 for each 5 epochs. Moreover, we set the $\alpha_J=1$, $\beta_J=1$, $\alpha_P=1$, $\beta_P=0.5$, $\lambda_J=1$, $\lambda_P=1$ and $\lambda_f=1$, $\lambda_b=0.5$ in our experiment. 

\section{Results}

\subsection{Evaluation Metrics}
We evaluate the generated JHMs \& PAFs and recovered video frames separately. 
\subsubsection{JHMs \& PAFs Evaluation Metrics}
We employed Percentage of Correct Keypoint (PCK) as JHMs and PAFs evaluation metrics due to the complexity to directly evaluate the heatmap and 2D vector field. Before computing PCK, we assembled JHMs and PAFs to keypoints and calculated their corresponding coordinates using algorithms given by Openpose. The PCK which is computed as Equation~\ref{PCK} reflects the distance between correct body keypoints and predicted keypoints.
\begin{equation}
PCK_i@\alpha=\frac{1}{H}\sum\limits^{H}_{h=1}\mathbb{I}\left(\frac{||p_{h,i}-g_{h,i}||}{diag_h}\leq\alpha\%\right)
\label{PCK}
\end{equation}
In Equation~\ref{PCK}, H is the total number of detected human instances, i is the index of body keypoint, ${diag}_h$ is the diagonal length of the instance bounding box, $\mathbb{I}(\cdot)$ is an indicator function equals to 1 if the situation in bracket is satisfied otherwise 0.
\subsubsection{Recovered Video Evaluation Metrics}
As in Equation~\ref{back}, the video generation network is trained with a fixed background label. Therefore, it is meaningless to evaluate the whole frame using conventional synthetic image evaluation metrics. To focus on the generated human body, we use Mask-RCNN \cite{he2017mask} which is well-trained on COCO dataset to create masks from raw video frames and recovered frames then calculate the intersection over union (IoU) between them. This idea is inherently similar to Inception Score (IS) \cite{mirza2014conditional} which utilizes other neural network to evaluate. This way, the well-trained Mask-RCNN can automatically evaluate the synthetic human body's quality and the IoU could efficiently reflect the localization and pose discrepancy between real and recovered video frames. Given the above, we denote the Mask-RCNN as $\mathcal{M}(\cdot)$ and the IoU could be computed as Equation~\ref{IoU}.
\begin{equation}
IoU@\alpha=\frac{1}{N}\sum\limits_{n=1}^N\mathbb{I}(\frac{\mathcal{M}(\mathbf{S})\cap\mathcal{M}(\mathbf{I})}{\mathcal{M}(\mathbf{S})\cup\mathcal{M}(\mathbf{I})}\leq\alpha\%)
\label{IoU}
\end{equation}
where $N$ is the total frame number and $\mathbb{I(\cdot)}$ is defined the same as Equation~\ref{PCK}.

\subsection{Pose Estimation Quality}
\begin{figure*}[t!]
\centering
\includegraphics[width=0.33\linewidth]{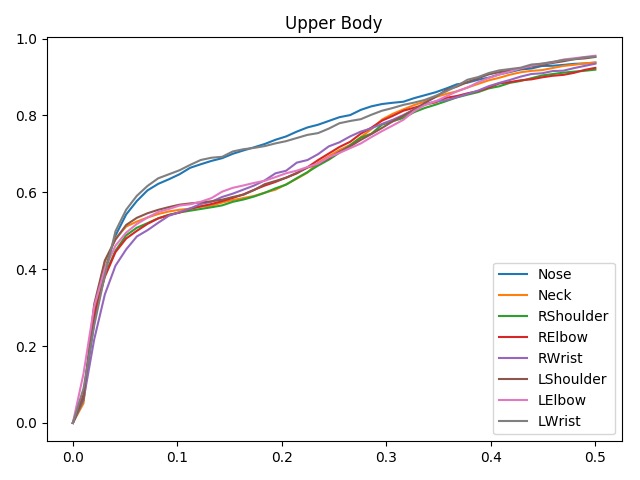}
\includegraphics[width=0.33\linewidth]{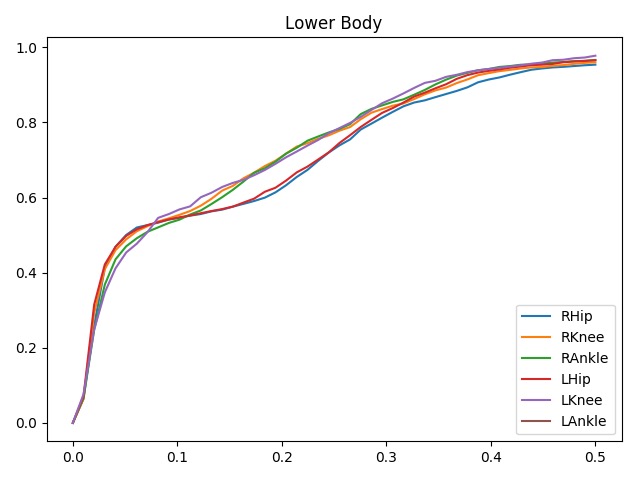}
\includegraphics[width=0.33\linewidth]{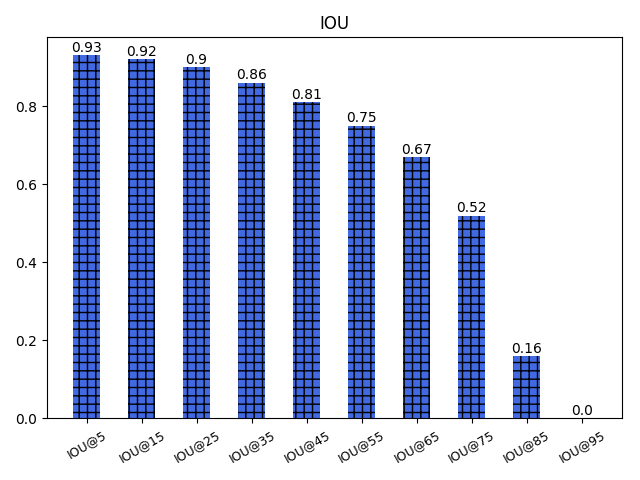}
\caption{PCK of detected body parts and IoUs of recovered frames. The left is the PCKs of Upper body. The middle is the PCKs of Lower body. The right is the IoUs of detected human instance in recovered frames.}
\label{pck}
\end{figure*}

Figure~\ref{pck} describes the PCKs of all 14 keypoints extracted from cross-modal mapping network. The mean PCK@0.2 of all 14 points are about 70\% which indicates the majority of keypoints will not diverge from ground truth away more than 20\% of diagonal length of instance's bounding box. This is an acceptable result since the spatial resolution of commercial Wireless signal is approximately 4cm. To further explore the gaps between vision-based approach, we presents the mPCK of both our method and Openpose on Table~\ref{cm}. The mean PCK@0.2 drops {10\%} by replacing source from image to wireless pixel. 

\subsection{Video Reconstruction Quality}
In Figure~\ref{sp}, we show our video generation results for single person scenario. The raw frames are recorded directly from the camera and the O-frames and C-frames are generated using JHMs and PAFs from Openpose and cross-modal network respectively. The O-frames have almost perfect quality except the face part. The reason for the blurred face perhaps because we just used an $256\times 128$ image as label. It could be further improved by using higher-resolution ground-truth. In comparison to the O-frames, the C-frames have a relatively low quality. The gap between O-frame and C-frame indicates that the wireless pixel cannot reflect the localization and movement as accurate as image pixels. However, the image quality of C-frame is quite enough to meet the requirements for behavior monitoring. And the image quality is supposed to be improved by training on more samples and manually annotating the ground-truth. 

\begin{figure}[t]
\centering
\begin{minipage}[b]{\columnwidth}
\includegraphics[width=0.49\linewidth, height=0.3\linewidth]{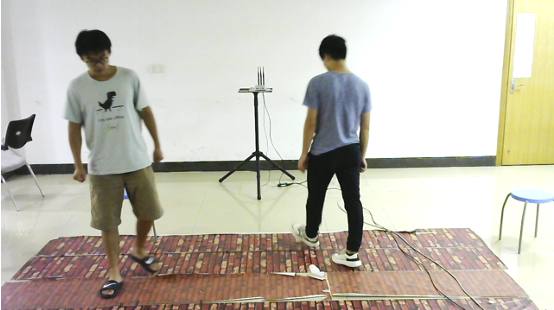}
\includegraphics[width=0.49\linewidth, height=0.3\linewidth]{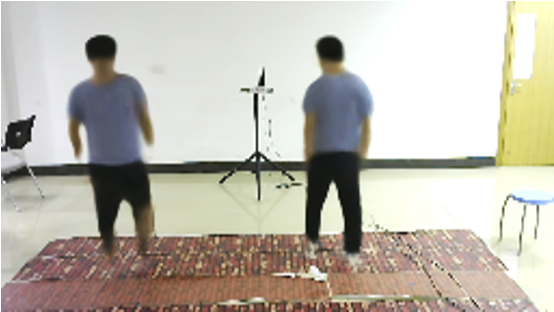}
\end{minipage}
\caption{A recovered frame example of the multiple-people scenario. The left is captured from camera and the right is generated from video generation network.}
\label{multiple people}
\end{figure}

Figure~\ref{multiple people} depicts an example of the multiple-people scenario. We can find that the image quality of it has no distinct difference from that of single person scenario while the appearances of generated human instances are almost random. This is because the network cannot match the human instance with their identities since the video frame only contains explicit information about who are in the RF perception field but cannot tell the identity of each one. We will address this issue in CSI2Video2 by obtaining human identity information using indoor identification and localization technology.

\begin{table}[h]
\caption{mIoU}
\label{cm}
\centering
\begin{tabular}{ccc}
\hline
Mask-RCNN&Person in WiFi&\bf{Ours}\\
\hline
0.79&0.68&0.66\\
\hline
\end{tabular}
\end{table}

Then, we illustrate the results of objective evaluation metrics. For this purpose, we compute the IoU scores as Equation~\ref{IoU} on different thresholds as depicted in Figure~\ref{pck}. We observe that the recovered frames have a high IoU score and decline gently at low threshold values. The IoU begins to decrease rapidly from 0.55 to 0.75. This indicates IoUs concentrates in this interval. Additionally, to our best knowledge, this is the first attempt to generate RGB surveillance video using WiFi signals. For this reason, we choose Person in WiFi \cite{wang2019person} and Mask-RCNN \cite{he2017mask} as baseline. Specifically, Person in WiFi is an end-to-end body segmentation and pose estimation method using WiFi signals. Mask-RCNN is a camera-based segmentation method. Since we utilized Mask-RCNN to generate annotations in the training phase, we still need new manually annotated labels for Mask-RCNN evaluation. To evaluate the results, we randomly selected 250 samples and manually annotated them. Table~\ref{cm} illustrates the performance between those three methods. Although those methods use completely different sources to generate masks, they are all toward fine-grained human body segmentation. The gap between video-based method is obvious. Incompletely reconstructed body part and overlapped body part may account for the gap. Although having an obvious gap compared to camera-based method, the IoU of masks from generated frames are comparable with that from WiFi signals directly. This indicates that the generated video does not lose information from WiFi signals, however, the wireless signals cannot contain as much information as video frame. Whilst this problem can be mitigated by adding more WiFi router pairs or more recorded OFDM subcarriers. 

\section{Conclusion}
In this paper, we propose CSI2Video, a novel human perception and video generation scheme that is able to generate RGB surveillance video in an accurate and real-time manner by conducting cross-modal mapping and video generation using the pervasive WiFi signals from commercial devices and a source of human identity information. Two tailored networks are designed to map CSI measurements to pose features and generate synthetic surveillance video respectively. To evaluate the generated video quality, we propose an innovative evaluation metric that can reflect the image quality and instance localization simultaneously. We conduct extensive experiments, including visualization and performance comparison, to demonstrate the effectiveness of the proposed system.

\bibliography{ref.bib}

\end{document}